\documentclass[runningheads]{llncs}
\pdfoutput=1
\usepackage{graphicx}
\usepackage{amsmath,amssymb}
\usepackage{color}
\usepackage[width=122mm,left=12mm,paperwidth=146mm,height=193mm,top=12mm,paperheight=217mm]{geometry}

\def\mP{\mathcal{P}}

\def\1n{\mathbf{1}_n}
\def\0{\mathbf{0}}
\def\1{\mathbf{1}}

\def\p{{\bf p}}

\def\tt{\mbox{\tiny $T$}}

\begin{document}
\pagestyle{headings}
\mainmatter
\title{The Curious Robot: Learning Visual Representations via Physical Interactions} 
\titlerunning{Learning Visual Representations via Physical Interactions}

\authorrunning{Pinto, Gandhi, Han, Park, Gupta}

\author{Lerrel Pinto,\  Dhiraj Gandhi,\  Yuanfeng Han,\  Yong-Lae Park, \\ Abhinav Gupta}
\institute{The Robotics Institute, Carnegie Mellon University}

\maketitle
\begin{abstract}
What is the right supervisory signal to train visual representations? Current approaches in computer vision use category labels from datasets such as ImageNet to train ConvNets. However, in case of biological agents, visual representation learning does not require millions of semantic labels. We argue that biological agents use physical interactions with the world to learn visual representations unlike current vision systems which just use passive observations (images and videos downloaded from web). For example, babies push objects, poke them, put them in their mouth and throw them to learn representations. Towards this goal, we build one of the first systems on a Baxter platform that pushes, pokes, grasps and observes objects in a tabletop environment. It uses four different types of physical interactions to collect more than 130K datapoints, with each datapoint providing supervision to a shared ConvNet architecture allowing us to learn visual representations. We show the quality of learned representations by observing neuron activations and performing nearest neighbor retrieval on this learned representation. Quantitatively, we evaluate our learned ConvNet on image classification tasks and show improvements compared to learning without external data. Finally, on the task of instance retrieval, our network outperforms the ImageNet network on recall@1 by 3\%.
\end{abstract}
\section{Introduction}
 Recently most computer vision systems have moved from using hand-designed features to feature learning paradigm. Much of the visual feature learning is done in a completely supervised manner using category labels. However, in case of biological agents, visual learning typically does not require categorical labels and happens in a ``unsupervised'' manner~\footnote{By ``unsupervised'' we mean no supervision from other agents but supervision can come from other modalities or from time}. 

Recently there has been a strong push to learn visual representations without using any category labels. Examples include using context from images~\cite{doersch2015}, different viewpoints from videos~\cite{wang2015}, ego-motion from videos~\cite{jayaraman2015learning} and generative models of images and videos~\cite{kingma2013auto,goodfellow2014generative,denton2015deep,radford2015unsupervised}. However, all these approaches still observe the visual world passively without any physical interaction with the world. On the other hand, we argue visual learning in humans (or any biological agent) require physical exploration of the world. Our babies play with objects: They push them, grasp them, put them in their mouth, throw them etc. and with every interaction they develop a better visual representation. 

In this paper, we break the traditional paradigm of visual learning from passive observation of data: images and videos on the web. We argue that the next step in visual learning requires signal/supervision from physical exploration of our world. We build a physical system (on a Baxter robot) with a parallel jaw gripper and a tactile skin-sensor which interacts with the world and develops a representation for visual understanding. Specifically, the robot tries to grasp objects, push objects, observe haptic data by touching them and also observe different viewpoints of objects. While there has been significant work in the vision and robotics community to develop vision algorithms for performing robotic tasks such as grasping, to the best of our knowledge this is the first effort that reverses the pipeline and uses robotic tasks for learning visual representations. 

\begin{figure*}[t!]
\begin{center}
\includegraphics[width=4.0in]{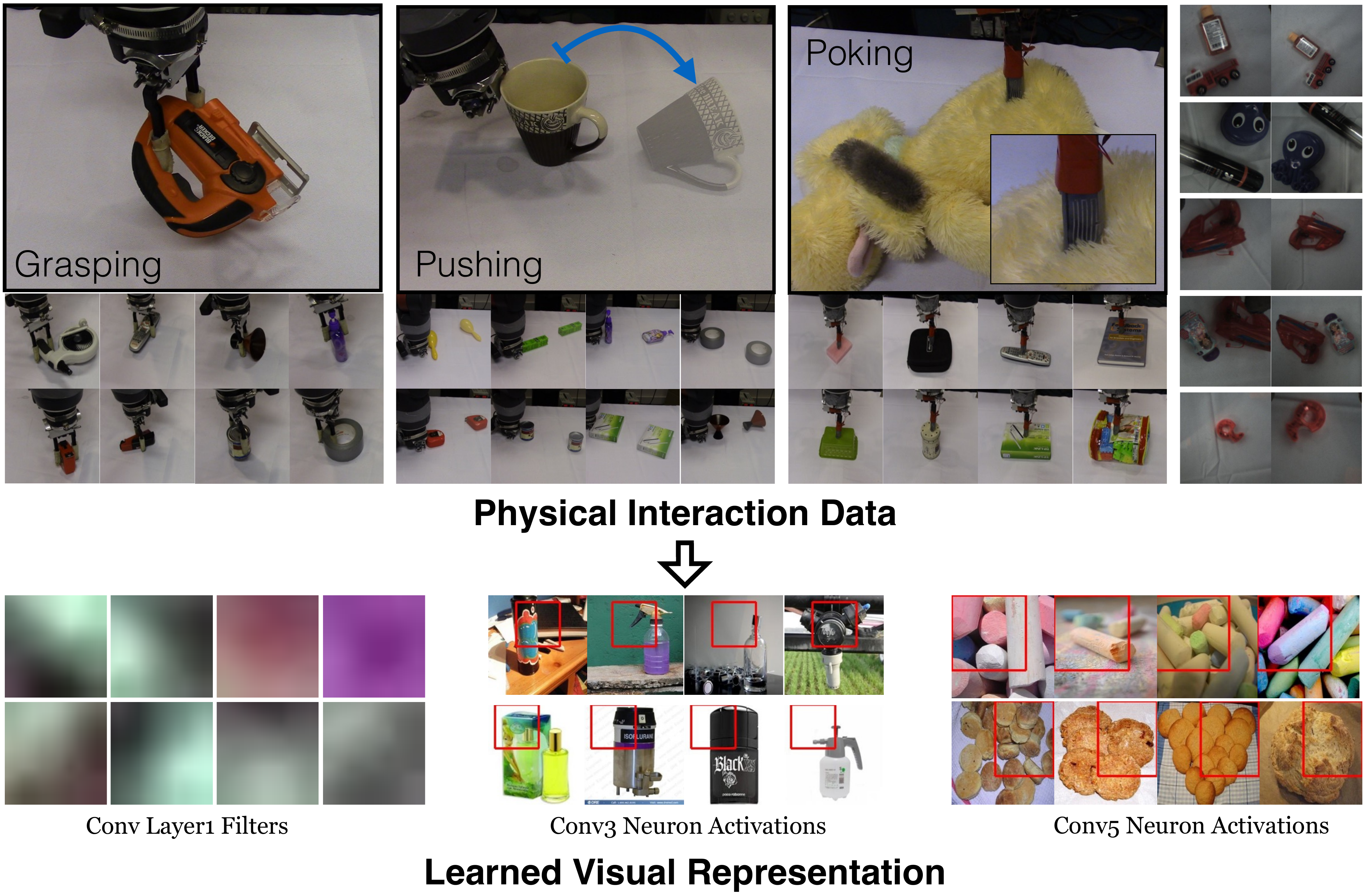}
\end{center}
\caption{Learning ConvNets from Physical Interactions: We propose a framework for training ConvNet using physical interaction data from robots. We first use a Baxter robot to grasp, push, poke and observe objects, with each interaction providing a training datapoint. We collect more than 130K datapoints to train a ConvNet. To the best of our knowledge, ours is one of the first system which trains visual representation using physical interactions.
}
\label{fig:intro}
\end{figure*}

We also propose a shared-ConvNet architecture where the first few convolutional layers are shared across different tasks, followed by 1-2 separate convolutional and fully connected layers for each task. Since our architecture is shared, every interaction creates a data point to train this ConvNet architecture. This ConvNet architecture then forms the learned visual representation. Our physical exploration data includes 40,287 grasps, 5,472 pushes, 1372 tactile sensing observations and 84,430 pairs of different viewpoints of the same object.

\section{Related Work}

Our work touches two threads: how do you learn a representation in an unsupervised way, and how do you interact with the world for learning.

\subsection{Unsupervised Learning}
Unsupervised learning of visual representations is one of the most challenging problems in computer vision. There are two common approaches to unsupervised learning: generative and discriminative. Early work in generative learning focused on learning visual representations that (a) can reconstruct images and (b) are sparse. This idea was extended by Hinton and Salakhutdinov~\cite{salakhutdinov2009deep}
to train a deep belief network in an unsupervised manner via stacking RBMs. Similar to this, Bengio et al. ~\cite{bengio2007greedy} investigated stacking of both RBMs and autoencoders.
Recently, there has also been a lot of success in the ability to generate realistic images. Some examples include the generative adversarial network (GAN)~\cite{goodfellow2014generative} framework and its variant including Laplacian GAN (LAPGAN)~\cite{denton2015deep}, DCGAN~\cite{radford2015unsupervised} and $S^{2}$-GAN~\cite{wang2016}. 

The discriminative approach to unsupervised learning is training a network on an auxiliary task where ground-truth is obtained automatically (or via sensors). For example, Doersch et al.~\cite{doersch2015} presented an approach to train networks via supervision from context. Other approaches such as~\cite{jayaraman2015learning,agrawal2015learning,mobahi2009deep} have tried to use videos and ego-motion to learn the underlying network. In an another work, \cite{wang2015} tracks patches in the video and uses the patches sampled from a track as different viewpoints of the same object instance and trains an embedding network. Finally, our approach is also related to other efforts which train deep networks using supervision from other sensors such as Kinect~\cite{eigen2014depth,wang2015designing} or motion information such as optical flow~\cite{walker2015dense}. 

However, all the above approaches still only observe passive data. We believe that the ability to physically interact with elements of the scene, is a key requirement for training visual representations. In fact, this has been supported by several psychological behavior experiments as well~\cite{held1963movement}.

\subsection{Robotic Tasks}
We note that most work in this domain has been in using vision to solve a task. Here, we use these tasks to learn good visual representations. We design our experiments in which our robot interacts with objects with four different types of interactions. 

\noindent {\bf Grasping.} Grasping is probably the oldest problem in the field  of robotics.  A  comprehensive  literature  review  of  this area  can  be  found  in ~\cite{bicchi2000robotic,bohg2014data}. Recently, data-driven learning-based approaches have started to appear, with initial work focused on using human annotators~\cite{lenz2013deep}. However, in this work we are more interested in building a self-supervision system~\cite{morales2004using,detry2009learning,Paolini_2014_7585,pinto2015supersizing,levine2016learning}. For the purpose of this work we use the dataset provided in ~\cite{pinto2015supersizing}.

\noindent {\bf Pushing.} The second task our system explores is the task of pushing objects. The origins of pushing as a manipulation task can be traced to the task of aligning objects to reduce pose uncertainty~\cite{balorda1990reducing,balorda1993automatic,lynch1996stable} and as a preceding realignment step before object manipulation ~\cite{dogar2011framework,yun1993object,zhou2016convex}. The implementation of pushing strategies in the mentioned work requires physics based models to simulate and predict the actions required to achieve a specific object goal state given an object start state. However, similar to grasping, in this work we are interested in self-learning systems which perform push interactions and use the sensor readings from robots as supervision.

\noindent {\bf Tactile Sensing.} The third task we explore is tactile sensing: that is, a robot with a skin sensor touches/pokes the objects followed by storing skin sensor readings. We then try to use the sensor reading to provide supervision to train the visual representations. Highly sensitive tactile optical sensors have been used for robot exoskeletal finger feedback control~\cite{park2009exoskeletal}. While there have been approaches that have attempted to combine tactile sensing with computer vision for object detection~\cite{schneider2009object}, this is the first paper that explores the idea of learning visual representation from tactile data. 

\noindent {\bf Identity Vision.} The final task we use our physical system for is to get multiple images of the same object from different viewpoints. This is similar to the idea of active vision~\cite{aloimonos1988active,wu20153d}. However, in most of these approaches the next best view is chosen after inference~\cite{wu20153d}. In this work, we sample thousands of such pairs to provide training examples for learning ConvNets.

\noindent {\bf Vision and Deep Learning for Robotics.} There has been a recent trend to use deep networks in robotics systems for grasp regression \cite{pinto2015supersizing,mahlerdex,redmon2014real} or learning policies for a variety of tasks ~\cite{levine2015learning,tzeng2015towards,finn2015deep}. In this paper, we explore the idea of using robotic tasks to train ConvNets as visual representation. We then explore the effectiveness of these for tasks such as object classification and retrieval. 
\section{\label{sec:formulation}Approach} 
We now describe the formulation of the four tasks: planar grasping, planar pushing, poking (tactile sensing) and identity vision for different viewpoints of the objects. 

\subsection{Planar Grasps} 
\begin{figure*}[t!]
\begin{center}
\includegraphics[width=4.0in]{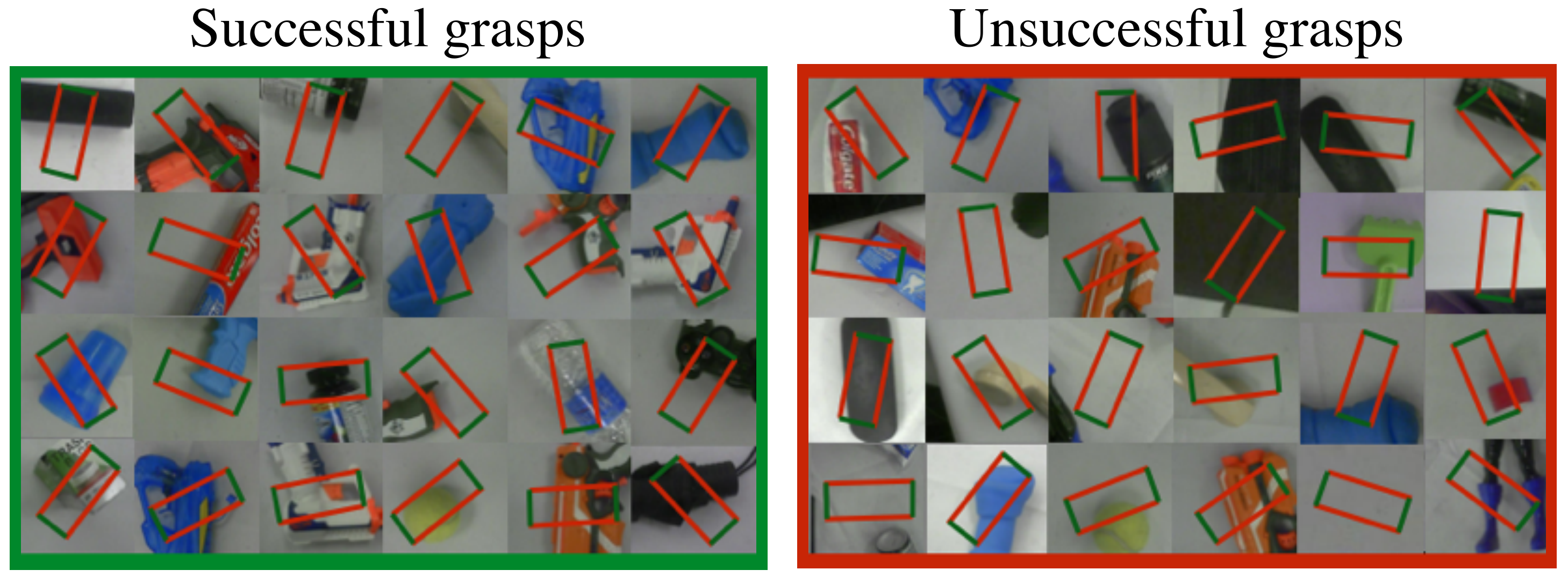}
\end{center}
\caption{Examples of successful (left) and unsuccesful grasps (right). We use a patch based representation: given an input patch we predict 18-dim vector which represents whether the center location of the patch is graspable at $0^{\circ}$, $10^{\circ}$, \dots $170^{\circ}$.}
\label{fig:grasp_data}
\end{figure*}

We use the grasp dataset described in our earlier work \cite{pinto2015supersizing} for our experiments on the grasping task. Here, the grasp configuration lies in 3 dimensions, $(x,y)$: position of grasp point on the surface of table and  $\theta$: angle of grasp. The training dataset contains around 37K failed grasp interactions and around 3K successful grasp interactions as the training set. For testing, around 2.8K failed and 0.2K successful grasps on novel objects are provided. Some of the positive and negative grasp examples are shown in Figure~\ref{fig:grasp_data}.

\subsubsection{Grasp prediction formulation:} 
The grasp prediction problem can be formulated as finding a successful grasp configuration $(x_S,y_S,\theta_S)$ given an image of an object $I$. However, as mentioned in \cite{pinto2015supersizing}, this formulation is problematic due to the presence of multiple grasp locations for each object and that ConvNets are significantly better at classification than regression to a structured output space. Given an image patch, we output an 18-dimensional likelihood vector where each dimension represent the likelihood of whether the center of the patch is graspable at $0^{\circ}$, $10^{\circ}$, \dots $170^{\circ}$. Therefore, the grasping problem can be thought of as 18 binary classification problems. Hence the evaluation criterion is binary classification i.e. given a patch and executed grasp angle in the test set, predict whether the object was grasped or not.

\subsection{Planar Push}
\begin{figure*}[t!]
\begin{center}
\includegraphics[width=4.0in]{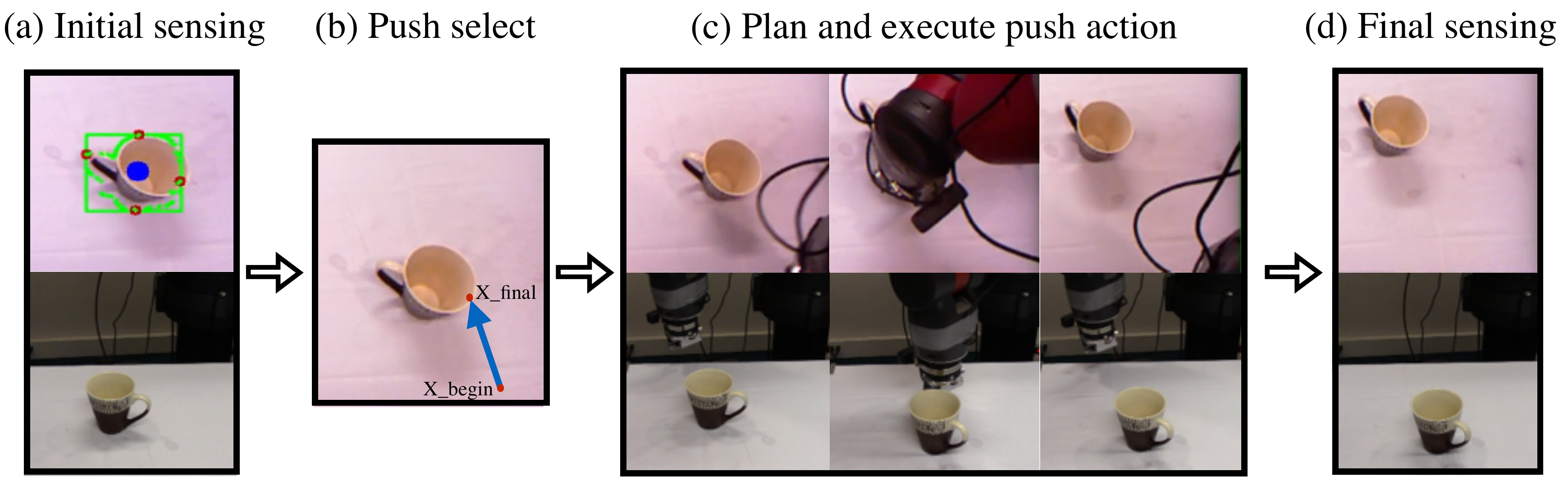}
\end{center}
\caption{We detect an object and apply planar push. Our push action is parameterized in terms of two 3D points: $X_{begin}$ and $X_{final}$.}
\label{fig:push_data_collection}
\end{figure*}

We use a Baxter robot to collect push data as described in Figure \ref{fig:push_data_collection}. Given an object placed on the robot's table and the model of the table, we perform background subtraction to determine the position of the object. After detecting an object using image $I_{begin}$, we sample two points on the workspace: the first point is the start point $X_{begin} = (x_{begin},y_{begin},z_{begin})$ from where the robot hand starts accelerating and gaining velocity. This start point is sampled from a Von Mises distribution to push the object towards the center of the robot's workspace.  We also sample another 3D point, $X_{final} = (x_{final},y_{final},z_{final})$. This point defines the location at which the robot hand stops accelerating. Note that: (a) $z_{begin} = z_{final}$ since we are dealing with planar pushing; (b) $X_{final}$ and $X_{begin}$ define both the direction and velocity/force with which the object is hit by the robot. Therefore, in this work, we parametrize push actions with 5 parameters: $(x_{start},y_{start},x_{final},y_{final},z_{pushHeight})$.

\begin{figure*}[t!]
\begin{center}
\includegraphics[width=4.0in]{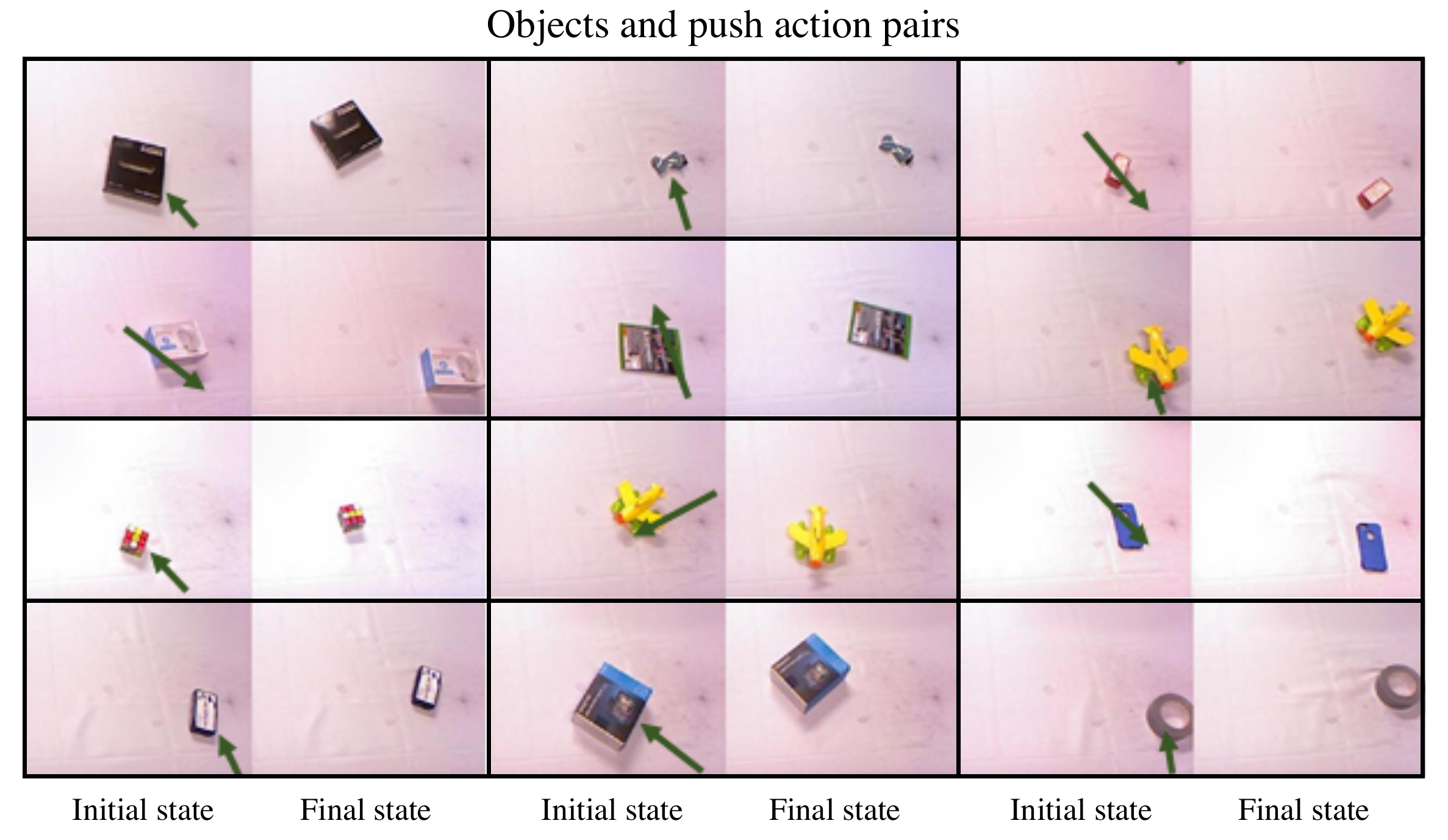}
\end{center}
\caption{Examples of initial state and final state images taken for the push action. The arrows demonstrate the direction and magnitude of the push action.}
\label{fig:push_data}
\end{figure*}

An off-the-shelf planner plans and executes the push action $X_{begin} \rightarrow X_{final}$. The arm then retracts back,and $I_{final}$ is recorded. We collect 5K push actions on 70 objects using the above described method. Some of these push actions are visualized in Figure \ref{fig:push_data}.

\subsubsection{Push prediction formulation:}
For incorporating learning from push, we build a siamese network (weights shared). One tower of siamese network takes $I_{begin}$ as input and the other tower takes $I_{end}$ as input. The output features of the two towers are then combined using fully connected layer to regress to what action caused this transformation. The push-action is parametrized by  $\{X_{begin},X_{final}\}$ and mean squared error is used as the regression loss. This action formulation captures the relevant magnitude as well as the localization and direction of the push.

\subsection{Poking: Tactile Sensing}
\begin{figure*}[t!]
\begin{center}
\includegraphics[width=4.0in]{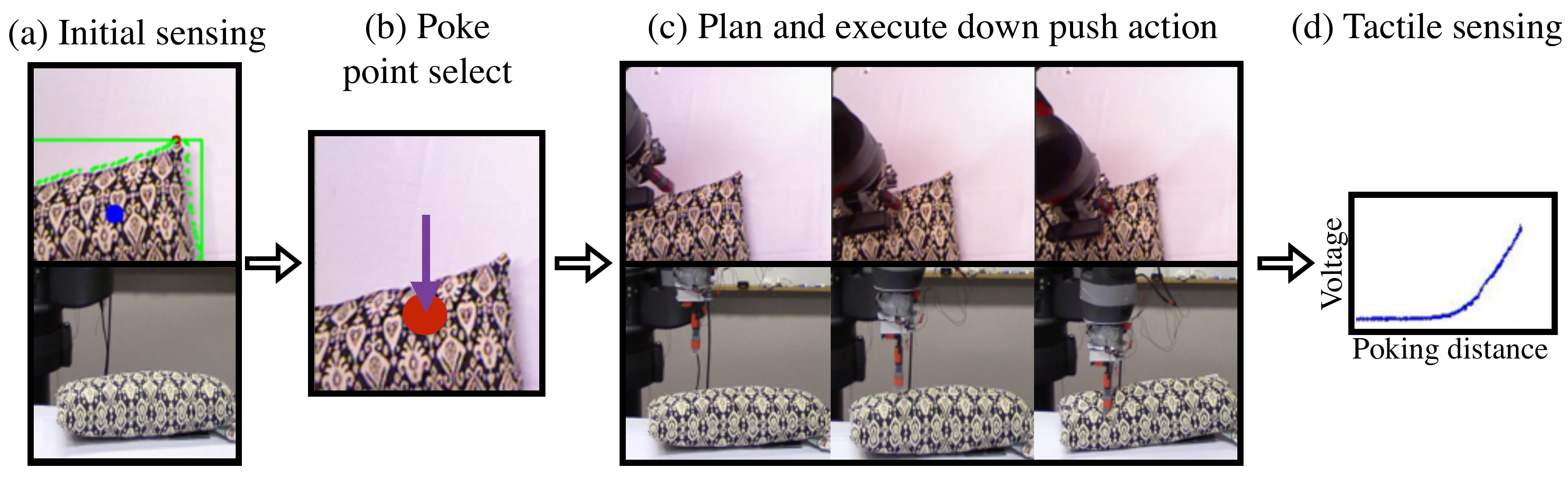}
\end{center}
\caption{The figure shows how the poke data with tactile sensing is collected. The profile of the tactile graph provides cues about material with which the object is made.
}
\label{fig:poke_data_collection}
\end{figure*}

In this task, the robot pushes the object vertically into the table until the pressure applied on the object exceeds a limit threshold (Figure \ref{fig:poke_data_collection}). A random point is sampled inside the object as the poke location. This pressure is sensed by robust tactile sensors attached on the finger of the robot and is continuously recorded. On reaching the limit threshold, the arm pulls away from the object and repeats the motion for around 10 times per object. A total of 1K pushes are collected on 100 diverse objects.

The tactile skin-sensor used in this work increases its electrical resistance monotonically on the application of pressure. Hence an increase in pressure correlates to an increase in resistance which then correlates to an increase in voltage drop across the sensor. This voltage drop $\p_{do}$ is sensed via an Arduino and logged with appropriate time stamps.

\begin{figure*}[ht]
\begin{center}
\includegraphics[width=4.3in]{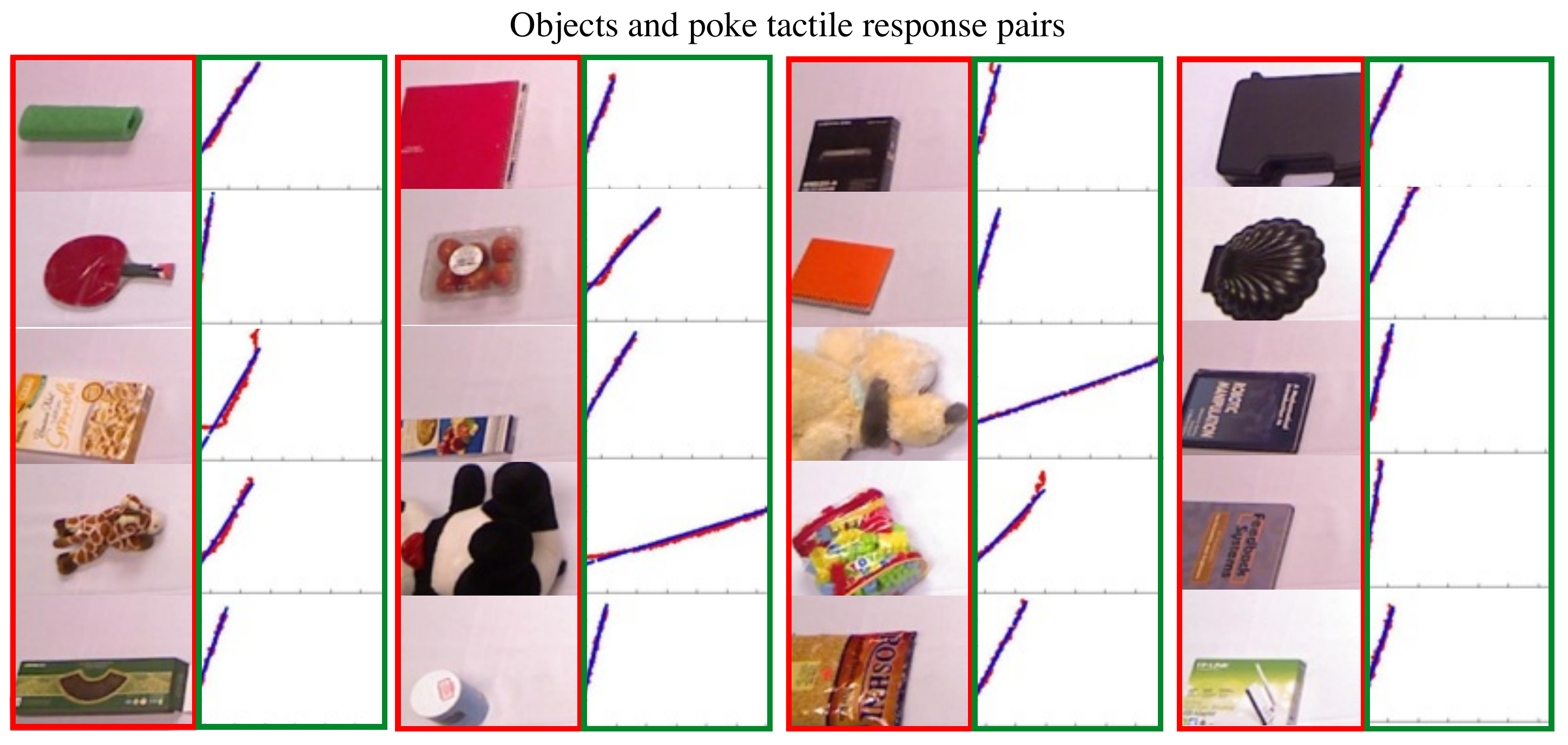}
\end{center}
\caption{Examples of the data collected by the poking action. On the left we show the object poked, and on the right we show force profiles as observed by the tactile sensor.}
\label{fig:poke_data}
\end{figure*}

During poking, the pressure voltage data stream $\p_{do}$ while pushing into the object is recorded. Some of this data are visualized in Figure \ref{fig:poke_data}. It can be noted how soft objects like the plush toy have a more gradual force response than harder objects like the hardcover book.  

\subsubsection{Tactile prediction formulation:}
The poke tactile prediction problem is formulated as a regression of the polynomial parametrization $\mP(\p_{do})$ of the poke action. Since a line parametrization works well to describe the poke response (Figure \ref{fig:poke_data}), we use a linear parametrization making the problem a regression to two values (the slope and the intercept). Therefore, given an image of the object our ConvNet predicts the slope and the intercept of $\mP(\p_{do})$.

\subsection{Identity Vision: Pose Invariance}
\begin{figure*}[h]
\begin{center}
\includegraphics[width=4.0in]{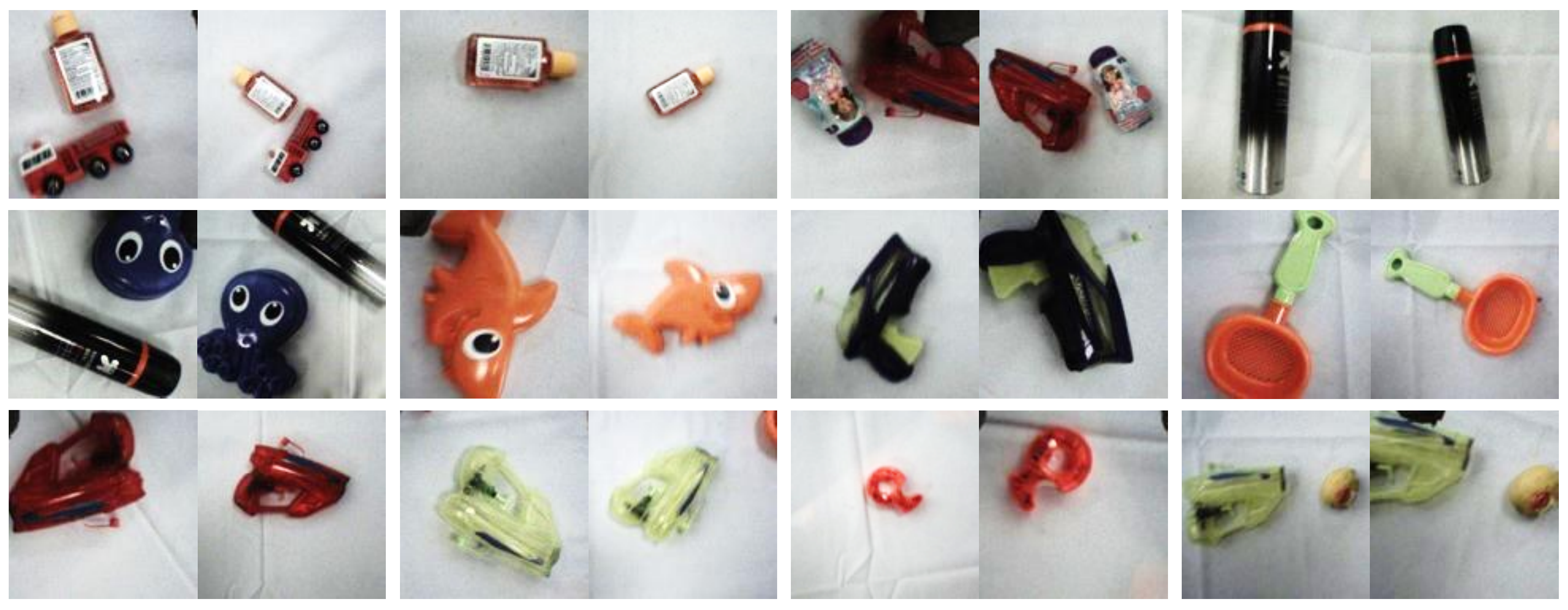}
\end{center}
\caption{Examples of objects in different poses provided to the embedding network.}
\label{fig:pose}
\end{figure*}

Given the grasping and pushing tasks described above, pairs of images in any one task's interaction contains images of objects with multiple viewpoints. The grasping dataset contains around 5 images of the object grasped from multiple viewpoints in every successful grasp interaction, while the planar push dataset contains 2 images of the object pushed in every push interaction. Figure~\ref{fig:pose} shows some examples of the pair of images of objects in different poses. In total, we use around 42K positive pairs of images and 42K negative pairs (images from different interactions).

\subsubsection{Pose invariance formulation}
The pose invariance task is meant to serve as a supervisory signal to enforce images of objects in the same task interaction to be closer in fc7 feature space. The problem is formulated as a feature embedding problem, where, given two images, the distance between the features should be small if the images are from the same robot interaction and large if the images are from different robot interactions.

\subsection{Network Architecture}
\begin{figure*}[t]
\begin{center}
\includegraphics[width=4.0in]{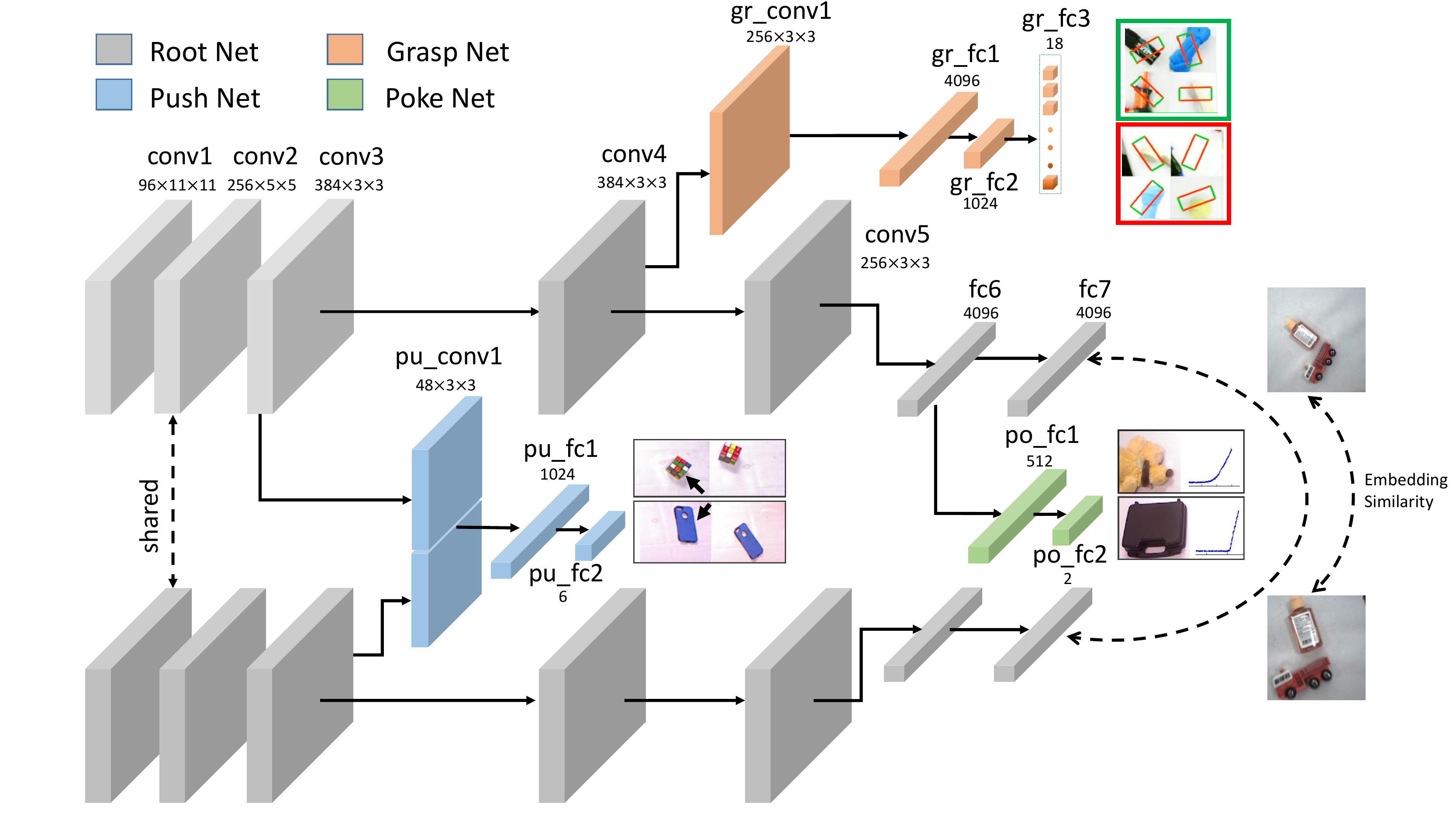}
\end{center}
\caption{Our shared convolutional architecture for four different tasks.}
\label{fig:Network}
\end{figure*}

We now describe our shared network architecture for the four tasks described above. The network architecture used is summarized in Figure \ref{fig:Network}. The network exploits the hierarchical sharing of features based on the complexity of the task to learn a common representation at the root network. Hence our network architecture can be seen as a root network that learns from every datapoint; this root network is augmented with specialized task networks emanating from various levels. This is based on the insight that tasks which require simpler representations like the push action prediction should be predicted lower in the network than more complex tasks like object embedding. Specifically, push action needs pose-variance. However, more complex tasks like object embedding require the learning of pose invariant feature representations, which are usually learnt in the higher layers.

The tasks we deal with in this work require either one image as input (like the grasp prediction task and the poke tactile task) or two images (like the push action prediction task, the action invariance task). We however would like to learn a single root network for all these tasks. For this purpose the root network in both the streams share parameters and are updated simultaneously. The gray blocks in Figure \ref{fig:Network} show the root network which updates its parameters using loss gradients backpropagated from every task.

\subsubsection{Root network:}
The root network follows the architecture of layer scheme of AlexNet ~\cite{krizhevsky2012imagenet} and can be seen as the gray network in Figure \ref{fig:Network}. The first convolutional layer (conv1) consists of 96 kernels with kernel size of $11\times11$. This convolutional layer and all the succeeding layers use a non linear Rectified Linear Unit (ReLU) as the transfer function. Local response normalization (LRN) is used and is followed by a spatial Max-Pooling (MP) of kernel size $3\times3$. This is followed by the second convolutional layer (conv2) that has 256 kernels of $5\times5$ kernel size and is followed by a LRN and a MP of $3\times3$. The third convolutional layer (conv3) has 386 $3\times3$ kernels which is followed by the fourth convolutional layer (conv4) with 384 $3\times3$ kernels. The fifth convolutional layer (conv5), with 256 $3\times3$ kernels, is followed by a MP with a $3\times3$ kernel. The convolutional layers are followed by two fully connected layers (fc6 and fc7) that have 4096 neurons each and are each followed by a ReLU. Since some tasks require two images as input, a clone of the root network is maintained with shared weights.

\noindent \textbf{Grasp network:} The input to the grasp network (orange blocks in Figure \ref{fig:Network}) emanates from the conv4 output of the root network. This network passes the input through one convolutional layer (gr\_conv1) with 256 kernels of $3\times3$ kernel size followed by a $3\times3$ MP. This then passes into a fully connected layer (gr\_fc1) with 4096 neurons followed by gr\_fc2 with 1024 and a final gr\_fc3 with $18\times2$ neurons.

The error criterion used is the 18-way binary classifier used in \cite{pinto2015supersizing}. Given a batch size $B$, with an input image $I_i$, the label corresponding to angle $\theta_i$ defined by $l_i\in \{0,1\}$ and the forward pass binary activations $A_{ji}$ on the angle bin $j$ \, the loss $L$ is:

\begin{equation}
L = \sum\limits_{i=1}^B\sum\limits_{j=1}^{N=18}\delta(j,\theta_i)\cdotp \textrm{softmax}(A_{ji},l_i)
\end{equation}

After an input training image $I_i$ is passed into the root network, the conv4 output is input into the grasp network that generates the classification loss that is backpropagated all through the chain. The weight update for the grasp network uses standard RMSprop, however the gradients for the root network are stored in memory and waits to be aggregated with the gradients from the other tasks.

\noindent \textbf{Push network:} The input to the push network (blue blocks in Figure \ref{fig:Network}) emanates from the conv3 output of the root network. This network consists of one convolutional layer (pu\_conv1) with 48 kernels of $3\times3$ kernel size, a fully connected layer (pu\_fc1) with 1024 neurons and a final fully connected layer (pu\_fc2) with 5 neurons corresponding to the size of the action prediction. Since the input for the push task is two images, the outputs from the pu\_conv1 for both the images are concatenated to result in a 96 dimensional feature map that is fed into pu\_fc1. A mean squared error (MSE) criterion is used as this task's loss.

Given input images $I_{begin}$ and $I_{final}$, $I_{begin}$ is passed in the root network while $I_{final}$ is passed through the clone of the root network. The conv3 outputs from both the networks pass through two copies of pu\_conv1 which is then concatenated and passed through the push network to generate MSE regression loss. The loss is backpropagated through the chain, with the weights in the push network getting updated in the batch using RMSprop while the gradients in the root network are stored in memory to be aggregated later. For the weights in pu\_conv1, the gradients are accumulated and mean-aggregated before an update.

\noindent \textbf{Poke network:} The input to the poke network (green in Figure \ref{fig:Network}) emanates from the fc6 of the root network. The poke network then passes the input into a fully connected layer po\_fc1 with 512 neurons followed by po\_fc2 with 2 neurons as the tactile prediction. The MSE criterion is used to provide the regression loss for this task.

Given an input training image, it is first passed through the root network. The fc6 output is then fed as an input to the poke network that makes a tactile prediction. The loss from the MSE criterion is backpropagated through the chain.


\noindent \textbf{Identity similarity embedding:} Feature embedding is done via a cosine embedding loss on the fc7 feature output. Given a pair of images, one is passed through the root network while the other is passed through a clone of the root network. The cosine embedding loss on the fc7 feature representations are then backpropagated through the chain and the gradients for the two copies are accumulated and mean aggregated.

\noindent \textbf{Training details:} We follow a two-step training procedure for the network. In the first stage, we first initialize the root network (upto conv4) and the grasp network with Gaussian initialization. We train only the grasp network and the lower root network for 20K iterations on the grasp data. 

In the second stage, we create the full architecture with first conv4 copied from the grasp learning. Then, batches with size of 128 are prepared for each of the 4 tasks and are sequentially input into the network. Weights for the grasp, push and poke networks are updated during their respective backward propagation cycles, while the gradients for the root and the clone network are accumulated until one batch for each of the tasks is complete. After one cycle of the 4 task batches, the accumulated gradients for the root and clone network are mean aggregated and a weight update step is taken. 


\section{Results}
We now demonstrate the effectiveness of learning visual representations via physical interactions. First, we analyze the learned network in terms of what it has learned and how effective the feature space is. Next, we evaluate the learned representations for tasks like image classification and image retrieval. Finally we analyze the importance of each task in the learnt representations using a task ablation analysis.

\subsection{Analyzing the ConvNet}
\begin{figure*}[t!]
\begin{center}
\includegraphics[width=4.0in]{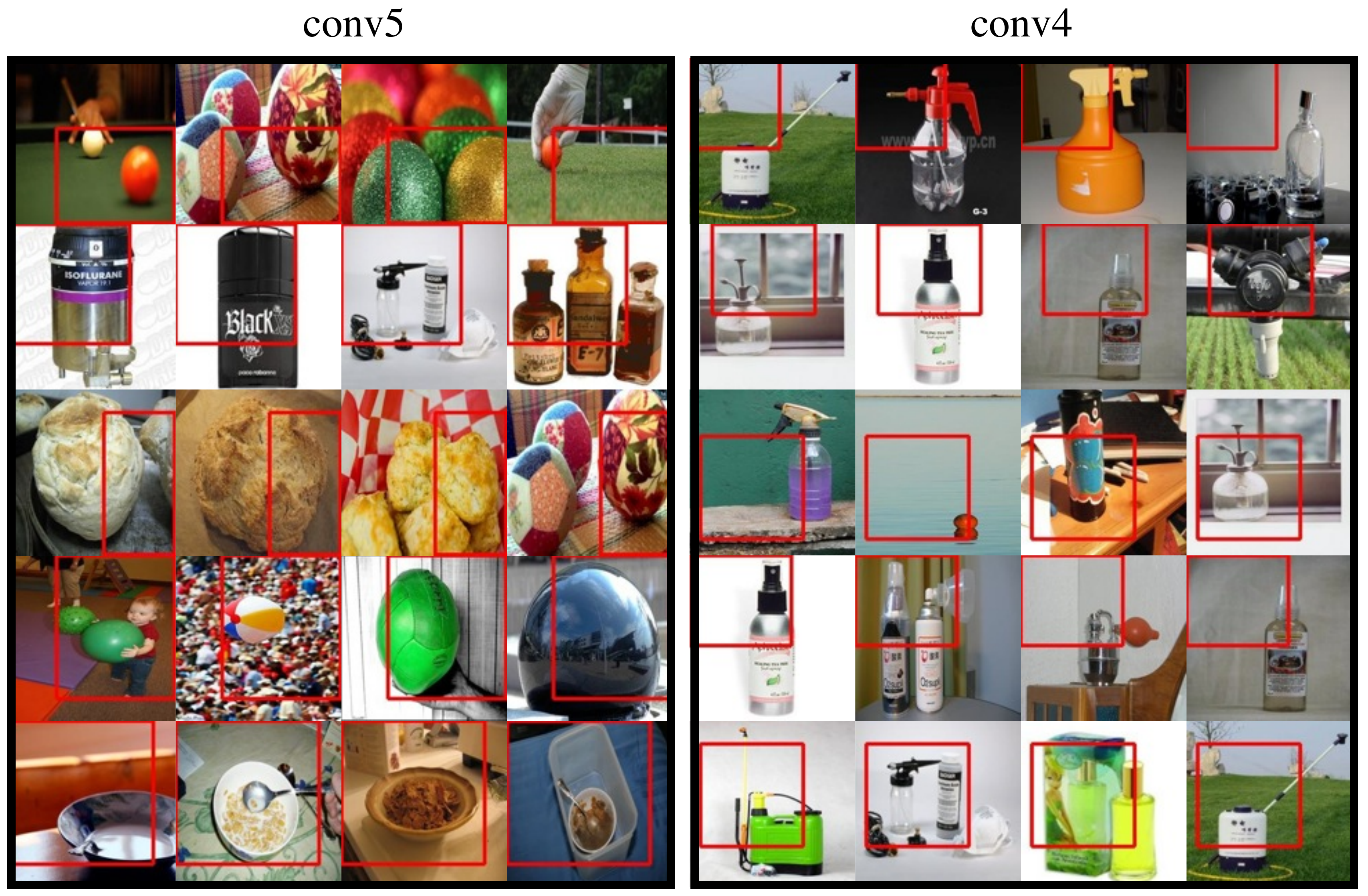}
\end{center}
\caption{The maximally activating images for the conv5 and conv4 neurons of our root network is shown here.}
\label{fig:neuron}
\end{figure*}

As a first experiment, we visualize the maximum activations of neurons in layer 4 and layer 5 of the network. Specifically, we run our learned network on 2500 ImageNet images corresponding to household items and find the images that maximally activates different neurons. We show these maximally activated images along with the receptive fields for some conv5 and conv4 neurons of our root network in Figure \ref{fig:neuron}. Notice that conv5 is able to correlate strong shape attributes like the spherical objects in row 4, the cereal bowls in row 5 and the circular biscuits in row 3. This is quite intuitive since tasks such as pushing, grasping etc. are functions of the object shapes.

\begin{figure*}[t!]
\begin{center}
\includegraphics[width=4.0in]{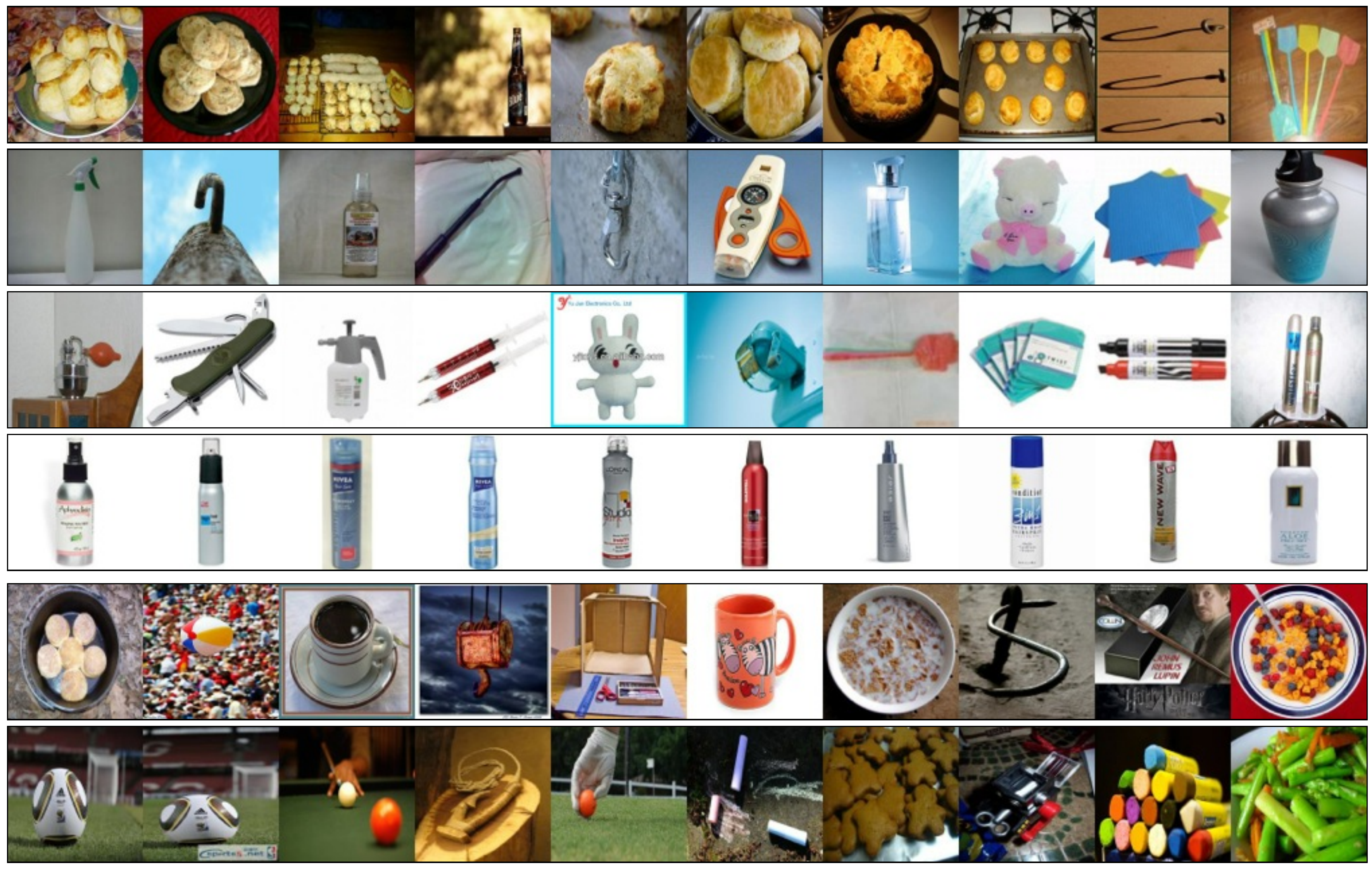}
\end{center}
\caption{The first column corresponds to query image and rest show the retrieval. Note how the network learns that cups and bowls are similar (row 5).}
\label{fig:NN_ImageNet}
\end{figure*}

As a next experiment, we analyze the learned network without fine tuning for the task of nearest neighbor. We use 25 household objects as query images and 2500 ImageNet images as the training dataset. We use the root network's conv5 feature space to perform nearest neighbors (See Figure \ref{fig:NN_ImageNet}). Again, as expected, the nearest neighbors seem to be based on shape attributes.

\subsection{Classification}
For analyzing the effectiveness of our learned visual representation, we would like to analyze the root network of our learnt robot task model on ImageNet ~\cite{deng2009imagenet} categories. However ImageNet image categories are of a wide variety while the image data the robot sees is mostly of objects on a tabletop setting. Hence for a more reasonable study we report results on a dataset containing 25 object synsets: atomizer, ball, biscuit, bomb, bottle, bowl, box, chalk, cup, device, hook, marker, pen, plush, rod, shaker, sharpener, sponge, spoons, spray, stapler, swatter, tool, toys and vegetable. Since some of these synsets contain very few images, 100 images from each of these are accumulated to make a 2500 ImageNet household dataset. We further evaluate classification on the UW RGBD dataset~\cite{lai2011large}, and the Caltech-256 dataset. Results of classification accuracy on these dataset can be found in Table \ref{tab:classification}. We also report the performance of the network trained only on identity data and an auto-encoder trained on all the data.

It is noteworthy that our network, finetuned on the ImageNet household 25 class dataset, gives around 35.4\% accuracy which is 10.4\% higher than training from scratch. This means that our network learns some features from the robot task training data that helps it perform better on the ImageNet task. This is quite encouraging since the correlation between robot tasks and semantic classification tasks have been assumed for decades but never been demonstrated. On the UW dataset, using leave one-out methodology, we report an accuracy of 69.3\%, which is about 22.5\% higher than learning from a scratch network. Similarly on the Caltech-256 dataset, our network has a performance of 31.7\% which is 7.5\% higher than learning from a scratch network.

\begin{table}[t]
\centering
\caption{Classification accuracy on ImageNet Household, UW RGBD and Caltech-256}
\label{tab:classification}
\setlength{\tabcolsep}{4pt}
\begin{tabular}{lccc}
                                           & \begin{tabular}[c]{@{}c@{}}Household\end{tabular} & \begin{tabular}[c]{@{}c@{}}UW RGBD\end{tabular} & Caltech-256 \\ \cline{2-4} 
Root network with random init.             & 0.250                                                        & 0.468                                                                & 0.242       \\
Root network trained on robot tasks (\textbf{ours}) & 0.354                                                        & 0.693                                                                & 0.317       \\
AlexNet trained on ImageNet                & 0.625                                                        & 0.820                                                                & 0.656       \\
Root network trained on identity data      & 0.315                                                        & 0.660                                                                & 0.252       \\
Auto-encoder trained on all robot data      & 0.296                                                        & 0.657                                                                & 0.280      
\end{tabular}
\end{table}

Similar to an unsupervised baseline~\cite{wang2015}, we train a network using 150K rotation and viewpoint data (identity data). Note that we use more datapoints than our multi task network in the paper. Yet, this performs worse than the network trained with robot tasks. Similarly, an auto-encoder trained on all the robot data perform worse than our network.

\subsection{Image Retrieval on UW RGBD dataset}
On the RGBD dataset~\cite{lai2011large} using fc7 features as visual representation, we perform and evaluate image retrieval. Our network's performance on instance level recall@k with k=1 and using cosine distance is 72\% which is higher than imageNet (69\%) and randomNet (6\%). On category level image retrieval ourNet's recall@1 is at 83\% a little lower than imageNet at 85\%. Table \ref{tab:recall} shows more recall@k analysis.

\begin{table}[]
\centering
\caption{Image Retrieval with Recall@k metric}
\label{tab:recall}
\setlength{\tabcolsep}{5pt}
\begin{tabular}{lcccc|cccc}
               & \multicolumn{4}{c|}{Instance level} & \multicolumn{4}{c}{Category level} \\ \cline{2-9} 
               & k=1     & k=5     & k=10   & k=20   & k=1     & k=5     & k=10   & k=20   \\ \cline{2-9}
Random Network & 0.062  & 0.219  & 0.331 & 0.475 & 0.150  & 0.466  & 0.652 & 0.800 \\
Our Network    & 0.720  & 0.831  & 0.875 & 0.909 & 0.833  & 0.918  & 0.946 & 0.966 \\
AlexNet        & 0.686  & 0.857  & 0.903 & 0.941 & 0.854  & 0.953  & 0.969 & 0.982
\end{tabular}
\end{table}



\subsection{Task Ablation Analysis}
To understand the contribution of each task to classification performance, we perform ablation analysis where we train our network excluding 1 out of 4 tasks (Table \ref{tab:ablation}). On all the three datasets, excluding Grasp data leads to the largest drop of performance which indicates that grasp task may be the most important among our tasks.

\begin{table}[h]
\centering
\caption{Task ablation analysis on classification tasks}
\label{tab:ablation}
\setlength{\tabcolsep}{5pt}
\begin{tabular}{lccc}

                                 & \begin{tabular}[c]{@{}c@{}}Household\end{tabular} & \begin{tabular}[c]{@{}c@{}}UW RGB-D\end{tabular} & Caltech-256                \\ \cline{2-4} 
All robot tasks                   & 0.354                                                        & 0.693                                           			& 0.317 \\
Except Grasp                    & 0.309                                                      & 0.632                                                                & 0.263                      \\ 
Except Push 			& 0.356                                   			& 0.710                                                                & 0.279                      \\ 
Except Poke                      & 0.342                                                       & 0.684                                                                & 0.289                      \\ 
Except Identity                   & 0.324                                   			& 0.711                                           			& 0.297                      \\ 
\end{tabular}
\end{table}

\section{Conclusion}
We present a method of learning visual representation using only robot interactions with the physical world. By experiencing over 130K physical interaction data points, our deep network is shown to have learnt a meaningful visual representation. We visualize the learned network, perform classification and retrieval tasks to validate our hypothesis. We note that this is just a small step in starting to integrate robotics and vision closely together. 
\medskip
\newline
\noindent{\textbf{Acknowledgement:} This work was supported by ONR MURI N000141612007, NSF IIS-1320083 and gift from Google. The authors would like to thank Yahoo! and Nvidia for the compute cluster and GPU donations respectively. The authors would also like to thank Xiaolong Wang for helpful discussions and code.}
\bibliographystyle{splncs}
\bibliography{egbib}
\end{document}